\documentclass[11pt,a4paper]{article}
\usepackage[hyperref]{acl2019}
\usepackage{nameref}
\usepackage{times}
\usepackage{latexsym}
\usepackage{natbib}
\usepackage{url}
\usepackage{booktabs}
\usepackage{tikz-qtree}
\usepackage{algpseudocode}
\usepackage{algorithm}
\usepackage{amsmath}
\usepackage{todonotes}
\usepackage{appendix}

\usepackage[utf8]{inputenc}

\aclfinalcopy 
\def\aclpaperid{1407} 

\usepackage{amsmath}

\DeclareMathOperator*{\argmin}{arg\,min}

\title{Correlating neural and symbolic representations of language}
\author{Grzegorz Chrupała \\
    Tilburg University\\
  \texttt{g.chrupala@uvt.nl} \\
  \And
  Afra Alishahi\\
      Tilburg University\\
  \texttt{a.alishahi@uvt.nl}}

\date{}

\begin{document}
\maketitle
\begin{abstract}
  Analysis methods which enable us to better understand the
  representations and functioning of neural models of language are
  increasingly needed as deep learning becomes the dominant approach
  in NLP. Here we present two methods based on Representational
  Similarity Analysis (RSA) and Tree Kernels (TK) which allow us to
  directly quantify how strongly the information encoded in neural
  activation patterns corresponds to information represented by
  symbolic structures such as syntax trees. We first validate our
  methods on the case of a simple synthetic language for arithmetic
  expressions with clearly defined syntax and semantics, and show that
  they exhibit the expected pattern of results. We then apply our methods to
  correlate neural representations of English sentences with their
  constituency parse trees.
\end{abstract}

\section{Introduction}
\label{sec:intro}
Analysis methods which allow us to better understand the
representations and functioning of neural models of language are
increasingly needed as deep learning becomes the dominant approach to
natural language processing. A popular technique for analyzing neural
representations involves predicting information of interest from the
activation patterns, typically using a simple predictive model such as
a linear classifier or regressor.  If the model is able to predict
this information with high accuracy, the inference is that the neural
representation encodes it.  We refer to these as {\it diagnostic models}.

One important limitation of this method of analysis is that it is only
easily applicable to relatively simple types of target information,
which are amenable to be predicted via linear regression or
classification. Should we wish to decode activation patterns into a
structured target such as a syntax tree, we would need to resort to
complex structure prediction algorithms, running the risk that the
analytic method becomes no simpler than the actual neural model.

Here we introduce an alternative approach based on correlating neural
representations of sentences and structured symbolic representations
commonly used in linguistics. Crucially, the correlation is in
similarity space rather than in the original representation space,
removing most constraints on the types of representations we can
use. Our approach is an extension of the Representational Similarity
Analysis (RSA) method, initially introduced by
\citet{kriegeskorte2008representational} in the context of
understanding neural activation patterns in human brains. 

In this
work we propose to apply  RSA to neural representations of strings from
a language on one side, and to structured symbolic representations of
these strings on the other side. To capture the similarities between
these symbolic representations, we use a {\it tree kernel}, a metric to
compute the proportion of common substructures between trees. This
approach enables straightforward comparison of neural and
symbolic-linguistic representations. Furthermore, we introduce
RSA$_{\textsc{regress}}$, a similarity-based analytic method which
combines features of RSA and of diagnostic models.

We validate both techniques on neural models which process a synthetic
language for arithmetic expressions with a simple syntax and semantics
and show that they behave as expected in this controlled setting. We
further apply our techniques to two neural models trained on English
text, Infersent \citep{conneau-etal-2017-supervised} and BERT
\citep{DBLP:journals/corr/abs-1810-04805}, and show that both models
encode a substantial amount of syntactic information compared to
random models and simple bag-of-words representations; we also show
that according to our metrics syntax is most salient in the
intermediate layers of BERT.



\section{Related work}
\label{sec:related}


\subsection{Analytic methods}
The dominance of deep learning models in NLP has brought an increasing
interest in techniques to analyze these models and gain insight into
how they encode linguistic information. For an overview of analysis
techniques, see \citet{belinkov2019analysis}.  The most widespread
family of techniques are {\it diagnostic models}, which use the
internal activations of neural networks trained on a particular task
as input to another predictive model. The success of such a predictive
model is then interpreted as evidence that the predicted information
has been encoded by the original neural model. The approach has also
been called {\it auxiliary task} \cite{adi2016fine}, {\it decoding}
\cite{alishahi2017encoding}, {\it diagnostic classifier}
\cite{hupkes2018visualisation} or {\it probing} \cite{P18-1198}.

Diagnostic models have used a range of predictive tasks, but since
their main purpose is to help us better understand the dynamics of a
complex model, they themselves need to be kept simple and
interpretable. This means that the predicted information in these
techniques is typically limited to simple class labels or values, as
opposed to symbolic, structured representations of interest to
linguists such as syntactic trees.
In order to work around this limitation \citet{tenney2018you} present
a method for probing complex structures via a formulation named {\it
  edge probing}, where classifiers are trained to predict various
lexical, syntactic and semantic relations between representation of
word spans within a sentence.

Another important consideration when analyzing neural encodings is the
fact that a randomly initialized network will often show non-random
activation patterns. The reason for this depends on each particular
case, but may involve the dynamics of the network itself as well as
features of the input data. For a discussion of this issue in the
context of diagnostic models see \citet{zhang2018language}.

Alternative approaches have been proposed to analyzing neural models
of language. For example, \citet{saphra2018understanding} train a
language model and parallel recurrent models for POS, semantic and topic tagging,
and measure the correlation between the neural representations of the
language model and the taggers.

Others modify the neural
architecture itself to make it more interpretable: \citet{W18-5403}
adapt layerwise relevance propagation \cite{bach2015pixel} to
Kernel-based Deep Architectures \cite{croce2017deep} in order
to retrieve examples which motivate model decisions. A vector
representation for a given structured symbolic input is built based on
kernel evaluations between the input and a
subset of training examples known as landmarks, and the network
decision is then traced back to the landmarks which had most influence
on it. In our work we also use kernels between symbolic structures,
but rather than building a particular interpretable model we propose a
general analytical framework.

\subsection{Representation Similarity Analysis}
\citet{kriegeskorte2008representational} present RSA as a variant of
pattern-information analysis, to be applied for
understanding neural activation patterns in human brains, for example
syntactic computations \citep{tyler2013syntactic} or
sensory cortical processing \citep{yamins2016using}.
The core idea is to find connections between data from neuroimaging,
behavioral experiments and computational modeling by correlating
representations of stimuli in each of these representation spaces via
their pairwise (dis)similarities.
RSA has also been used for measuring similarities between neural-network
representation spaces
\citep[e.g.][]{bouchacourt-baroni-2018-agents,chrupala2018symbolic}.

\subsection{Tree kernels}
For extending RSA to a structured representation space, we need a
metric for measuring (dis)similarity between two structured
representations. Kernels provide a suitable framework for this
purpose: \citet{collins2002convolution} introduce convolutional
kernels for syntactic parse trees as a metric which quantifies
similarity between trees as the number of overlapping tree fragments
between them, and introduce a polynomial time algorithm to compute
these kernels; \citet{moschitti2006making} propose an efficient
algorithm for computing tree kernels in linear average running time.

\subsection{Synthetic languages}
When developing techniques for analyzing neural network models of
language, several studies have used synthetic data from artificial
languages.  Using synthetic language has the advantage that its
structure is well-understood and the complexity of the language and
the statistical characteristics of the generated data can be carefully
controlled. The tradition goes back to the first generation of
connectionist models of language
\cite{elman1990finding,hochreiter1997long}. More recently,
\citet{W18-5414} and \citet{W18-5425} both use context-free grammars to
generate data, and train RNN-based models to identify matching numbers
of opening and closing brackets (so called Dyck languages).  The task
can be learned, but \citet{W18-5414} report that the models fail to
generalize to longer sentences.  \citet{W18-5456} also show that with
extensive training and the appropriate curriculum, LSTMs trained on
synthetic language can learn compositional interpretation rules.

Nested arithmetic languages are also appealing choices since they have
an unambiguous hierarchical structure and a clear compositional
semantic interpretation (i.e.\ the value of the arithmetic
expression).  \citet{hupkes2018visualisation} train RNNs to calculate
the value of such expressions and show that they perform and
generalize well to unseen strings.
They apply diagnostic classifiers to analyze the strategy employed by
the RNN model.

\section{Similarity-based analytical methods}
\label{sec:rsa}

RSA finds connections between data from two different representation
spaces. Specifically, for each representation type we
compute a matrix of similarities between pairs of stimuli. Pairs of these matrices are
then subject to second-order analysis by extracting their upper
triangulars and computing a correlation coefficient between them.

Thus for a set of objects $X$, given a similarity function $s_k$ for a
representation $k$, the function $S_k$ which computes the
representational similarity matrix is defined as:

\begin{equation} \label{eq:S_k}
\begin{split}
S_k(X) & = \mathbf{U} \\
U_{i,j}   & = s_k(X_i, X_j),
\end{split}
\end{equation}
and the RSA score between representations $k$ and $l$ for data
$X$ is the correlation (such as Pearson's correlation coefficient
$r$) between the upper triangulars $S_k(X)$ and $S_l(X)$, excluding
the diagonals.

\paragraph{Structured RSA} 
We apply RSA to neural representations of strings from a language on
one side, and to structured symbolic representations of these strings
on the other side.  The structural properties are captured by defining
appropriate similarity functions for these symbolic representations;
we use tree kernels for this purpose.

A tree kernel measures the similarity between a pair of tree
structures by computing the number of tree fragments they
share. \citet{collins2002convolution} introduce an algorithm for
efficiently computing this quantity;
a tree fragment in their formulation is a set of
connected nodes subject to the constraint that only complete
production rules are included.

\begin{figure}[htb]
  \centering
  \includegraphics[scale=0.5]{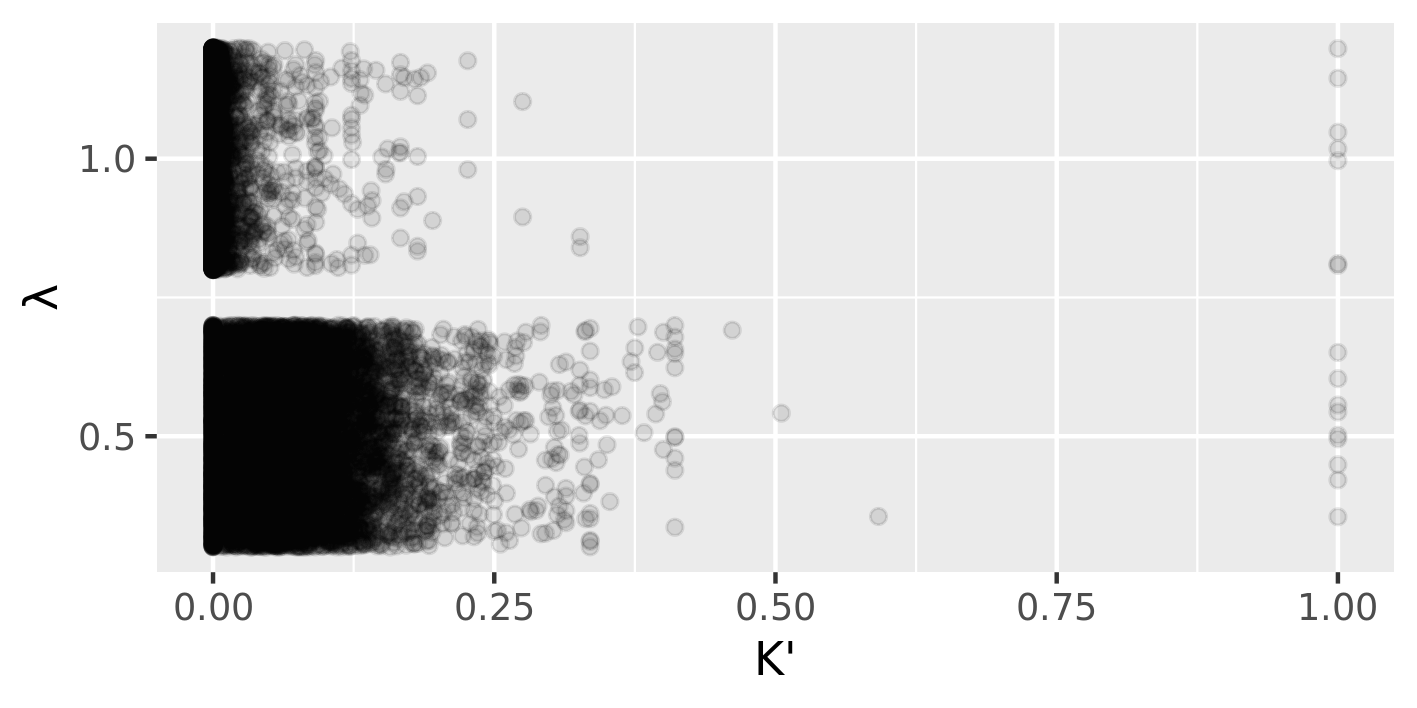}
  \caption{Distribution of values of the tree kernel for two settings
    of discounting parameter $\lambda$, for syntax trees of a sample
    of English sentences.}
  \label{fig:sim_distribution}
\end{figure}

Following \citet{collins2002convolution}, we calculate the tree kernel
between two trees $T_1$ and $T_2$ as:

    \begin{equation}
      K(T_1, T_2) = \sum_{n_1\in T_1}\sum_{n_2\in T_2} C(n_1, n_2, \lambda),
    \end{equation}
    where $n_1$ and $n_2$ are the complete sets of tree fragments in
    $T_1$ and $T_2$, respectively, and the function $ C(n_1, n_2,
    \lambda)$ is calculated as shown in figure \ref{fig:kernel_formula}.
    The parameter $\lambda$ is used to scale the relative importance of
    tree fragments with their size.
    Lower values of this parameter discount larger tree fragments
    in the computation of the kernel; the value 1 does not do any
    discounting.
   See Figure~\ref{fig:sim_distribution} for the illustration of the effect of the value of $\lambda$ on the kernel.

\begin{figure*}
    \begin{equation*}
      C(n_1, n_2, \lambda) =
      \begin{cases}
        0, & \mbox{if } \mathrm{prod}(n_1)\ne\mathrm{prod}(n_2) \\
        \lambda, & \mbox{if } \mathrm{preterm}(n_1) \land \mathrm{preterm}(n_2) \\
        \lambda  \prod_{i=1}^{\mathrm{nc}(n_1)} (1 + C(\mathrm{ch}(n_1, i), \mathrm{ch}(n_2, i), \lambda)) & \mbox{otherwise}.
      \end{cases}
    \end{equation*}
    \caption{Dynamic programming formula for computing a 
      convolution kernel, after \citet{collins2002convolution}. Here $\mathrm{nc}(n)$ is the number of children of a given
    (sub)tree, and $\mathrm{ch}(n, i)$ is its i$^{th}$
    child; $\mathrm{prod}(n)$ is the production of node $n$, and
    $\mathrm{preterm}(n)$ is true if $n$ is a preterminal node. }
      \label{fig:kernel_formula}
\end{figure*}
We work with normalized kernels: 
given a function $K$ which computes the raw count of tree fragments in
common between trees $t_1$ and $t_2$, the normalized tree kernel is
defined as:

\begin{equation}
    K'(t_1, t_2) = \frac{K(t_1, t_2)}{ \sqrt{K(t_1, t_1) K(t_2, t_2)} }.
\end{equation}

\begin{figure}[tbh]
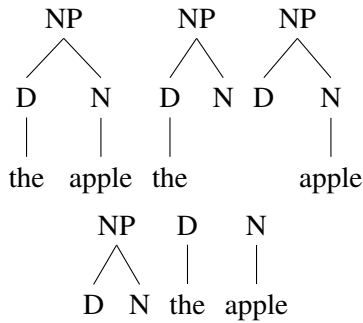

    \centering
    \Tree [.NP [.D the ] [.N apple ] ] 
    \Tree [.NP [.D the ] N ]
    \Tree [.NP D [.N apple ] ] \\
    \Tree [.NP D N ] 
    \Tree [.D the ] 
    \Tree [.N apple ] 
    \caption{The complete set of tree fragments as defined by the
      tree kernel for the syntax tree corresponding to {\it the apple}, after \citet{collins2002convolution}. }
    \label{fig:tree-fragments}
\end{figure}

Figure~\ref{fig:tree-fragments} shows the complete 
set of tree fragments which the tree kernel implicitly computes 
for an example syntax tree. 
    
\paragraph{RSA$_{\textsc{regress}}$}
Basic RSA measures correlation between similarities in two different
representations globally,
i.e.\ how close they are in their totality.  In contrast, diagnostic
models answer a more specific question: to what extent a particular
type of information can be extracted from a given representation. For
example, while for a particular neural encoding of sentences it may be
possible to predict the length of the sentence with high accuracy, the
RSA between this representation and the strings represented only by
their length may be relatively small in magnitude, since the neural
representation may be encoding many other aspects of the input in
addition to its length.

We introduce RSA$_{\textsc{regress}}$, a method which shares
features of both classic RSA as well as the diagnostic model
approach. Like RSA it is based on two similarity functions $s_k$ and
$s_l$ specific to two different representations $k$ and $l$. But
rather than computing the square matrices $S_k(X)$ and $S_l(X)$ for a
set of objects $X$, we sample a reference set of objects $R$ to act as
anchor points, and then embed the objects of interest $X$ in the
representation space $k$ via the representational similarity function
$\sigma_k$ defined as:\footnote{Note that $\sigma_k$ is simply a generalization of $S_k$ to the non-square case, namely $S_k(X) = \sigma_k(X, X)$.}
\begin{equation}\label{eq:sigma_k}
\begin{split}
    \sigma_k(X, R) & = \mathbf{V} \\
    V_{i, j}    & = s_k(X_i, R_j),
\end{split}    
\end{equation}
Likewise for representation $l$, we calculate $\sigma_l$ for the same
set of objects $X$. The rows of the two resulting matrices contain two
different views of the objects of interest, where the dimensions of
each view indicate the degree of similarity for a particular reference
anchor point. We can now fit a multivariate linear regression model to
map between the two views:

\begin{equation}
    \label{eq:rsa_regress}
 \widehat{\mathbf{B,a}} = \argmin_{\mathbf{B,a}} \mathrm{MSE}(\mathbf{B}\sigma_k(X, R) + \mathbf{a}, \sigma_l(X, R))
\end{equation}
\noindent
where $k$ is the source and $l$ is the target view, and $\mathrm{MSE}$
is the mean squared error.  The success of this model can
be seen as an indication of how predictable representation $l$ is from
representation $k$. Specifically, we use a cross-validated Pearson's
correlation between predicted and true targets for an $L_2$-penalized
model.

\section{Synthetic language}
\label{sec:synthetic}

Evaluation of analysis methods for neural network models is an open
problem.  One frequently resorts to largely qualitative evaluation:
checking whether the conclusions reached via a particular approach
have face validity and match pre-existing intuitions. However
pre-existing intuitions are often not reliable when it comes to
complex neural models applied to also very complex natural language
data. It is helpful to simplify one part of the overall system and
apply the analytic technique of interest on a neural model which
processes a simple and well-understood synthetic language.
As our first case study, we use a simple language of arithmetic
expressions. Here we first describe the language and its syntax and
semantics, and then introduce neural recurrent models which process
these expressions.

\subsection{Arithmetic expressions}
\label{sec:expressions}
Our language consists of expressions which encode addition and
subtraction modulo 10.  Consider the example expression {\tt
  ((6+2)-(3+7))}. In order to evaluate the whole expression, each
parenthesized sub-expression is evaluated modulo 10: in this case the
left sub-expression evaluates to 8, the right one to 0 and the whole
expression to 8. Table~\ref{tab:expressions} gives the context-free
grammar which generates this language, and the rules for semantic
evaluation. Figure~\ref{fig:expression-tree} shows the syntax tree for
the example expression according to this grammar. This language lacks
ambiguity, has a small vocabulary (14 symbols) and simple
semantics, while at the same time requiring the processing of
hierarchical structure to evaluate its expressions.\footnote{The
  grammar is more complex than strictly needed in order to facilitate the
  computation of the Tree Kernel, which assumes each vocabulary symbol
  is expanded from a pre-terminal node.}

\begin{table}
{\small
\begin{tabular}{ll}
  Syntax                                       & Meaning \\\toprule
  $E \rightarrow L~E_1~O~E_2~R $  & $[E] = [O]([E_1], [E_2]) $\\
  $E \rightarrow D            $  & $[E] = [D]$ \\
  $O \rightarrow +            $  & $[O] = \lambda x,y.x+y \textrm{ mod } 10 $\\
  $O \rightarrow -            $  & $[O] = \lambda x,y.x-y \textrm{ mod } 10 $\\
  $L \rightarrow \textrm{( } $   & \\
  $R \rightarrow \textrm{) } $   & \\
  $D \rightarrow 0 $             & $[D] = 0$\\
  \vdots & \vdots\\
  $D \rightarrow 9 $             & $[D] = 9$\\
\end{tabular}
 }
\caption{Grammar $G(L)$ of a language $L$ expressing addition and
  subtraction modulo 10 in infix notation. The notation $[\cdot]$
  stands for the semantic evaluation function. Subscripts on symbols
  serve to distinguish their multiple occurrence.}
\label{tab:expressions}
\end{table}

\begin{figure}[h]
    \centering
    {\small
      \Tree [.E [.L ( ] [.E [.L ( ] [.E [.D 6 ] ] [.O + ]  [.E [.D 2 ]
      ] [.R ) ] ] [.O - ]
            [.E [.L ( ] [.E [.D 3 ] ] [. O + ]  [.E [.D 7 ] ] [.R ) ] ] [.R ) ] ]
    }
    \caption{Syntax tree of the expression {\tt ((6+2)-(3+7))}.}
    \label{fig:expression-tree}
\end{figure}

\paragraph{Generating expressions}
In order to  generate expressions in $L$  we use the recursive function {\sc
  Generate} defined in Algorithm~\ref{algo:generate}. The function
receives two input parameters: the branching probability $p$ and the
decay factor $d$. In the recursive call to {\sc Generate} in lines
\ref{algo:line:rec1} and \ref{algo:line:rec2} the probability $p$ is
divided by the decay factor. Larger values of $d$ lead to the
generation of smaller expressions. Within the branching path in line
\ref{algo:line:op} the operator is selected
uniformly at random, and likewise in the non-branching path in line
\ref{algo:line:digit} the digit is sampled uniformly.

\begin{algorithm}
{\small
\begin{algorithmic}[1]
\Function{Generate}{$p$, $d$}
    \State branch $\sim$ \Call{Bernoulli}{$p$}
    \If { branch } 
        \State $e_1$ $\gets$ \Call{Generate}{$p/d$, $d$} \label{algo:line:rec1}
        \State $e_2$ $\gets$ \Call{Generate}{$p/d$, $d$} \label{algo:line:rec2}
        \State op $\sim$ \Call{Uniform}{$[+,-]$} \label{algo:line:op}
        \State \Return [$_{E}$ [$_{L}$ ( ] $e_1$ [$_{O}$ op ] $e_2$
        [$_{R}$ ) ] ]
    \Else 
        \State digit $\sim $ \Call{Uniform}{$[0,\ldots,9]$} \label{algo:line:digit}
        \State \Return  [$_{E}$ [$_{D}$ digit ] ] 
    \EndIf
\EndFunction
\end{algorithmic}
}
\caption{Recursive function for generating an expression of language
  $L$.}
\label{algo:generate}
\end{algorithm}

\subsection{Neural models of arithmetic expressions}
\label{sec:synthetic:neural}

We define three recurrent models which process the arithmetic
expressions from language $L$. Each of them is trained to predict a
different target, related either to the syntax of the language or to
its semantics. We use these models as a testbed for validating our
analytical approaches.  All these models share the same recurrent
encoder architecture, based on LSTM \citep{hochreiter1997long}.

\paragraph{Encoder}
The encoder consists of a trainable embedding lookup table for the
input symbols, and a single-layer LSTM. The state of the hidden layer
of the LSTM at the last step in the sequence is used as a
representation of the input expression.

\paragraph{\sc Semantic evaluation}
This model consists of the encoder as described above, which passes
its representation of the input to a multi-layer perceptron component
with a single output neuron. It is trained to predict the value of the input
expression, with mean squared error as the loss function.
In order to perform this task we would expect that the model needs to
encode the hierarchical structure of the expression to some extent while also
encoding the result of actually carrying out the operations of
semantic evaluation.

\paragraph{\sc Tree depth}
This model is similar to {\sc semantic evaluation} but is trained to
predict the depth of the syntax tree corresponding to the expression
instead of its value.
We expect this model to need to encode a fair amount of hierarchical
information, but it can completely ignore the semantics of the
language, including the identity of the digit symbols.

\paragraph{\sc Infix-to-prefix}
This model uses the encoder to create a representation of the input
expression, which it then decodes in its prefix form. For example, the
expression {\tt ((6+2)-(3+7))} is converted to {\tt
  (-(+62)(+37))}. The decoder is an LSTM trained as a conditional
language model, i.e.\ its initial hidden state is the output of the
encoder and its input at each step is the embedding of previous output
symbol. The loss function is categorical cross-entropy.
We would expect this model to encode the hierarchical structure in
some form as well as the identity of the digit symbols, but it can
ignore the compositional semantics of the language.

\subsection{Reference representations}

We use RSA to correlate the neural encoders from
Section~\ref{sec:synthetic:neural} with reference syntactic and
semantic information about the arithmetic expressions.
For the neural
representations we use cosine distance as the dissimilarity
metric. The reference representations and their associated
dissimilarity metrics are described below.

\paragraph{Semantic value}
This is simply the value to which each expression evaluates, also used
as the target of the {\sc semantic evaluation} model. As a measure of
dissimilarity we use the absolute difference between values, which ranges from 0 to 9.

\paragraph{Tree depth}
This is the depth of the syntax tree for each expression, also used as
the target of the {\sc tree depth} model. We use the absolute
difference as the dissimilarity measure. The dissimilarity is minimum 0 
and has no upper bound, but in our data the typical maximum value is
around 7.

\paragraph{Tree kernel}
This is an estimate of similarity between two syntax trees based on
the number of tree fragments they share, as described in Section~\ref{sec:rsa}. 
The normalized tree kernel metric ranges between 0 and 1, which we convert
to dissimilarity by subtracting it from 1. 

The semantic value and tree depth correlates are easy to
investigate with a variety of analytic methods including diagnostic
models; we include them in our experiments as a point of comparison.
We use the tree kernel representation to evaluate structured RSA for a
simple synthetic language.

\subsection{Experimental settings}
\begin{figure}[thb]
    \centering
    \includegraphics[scale=0.35]{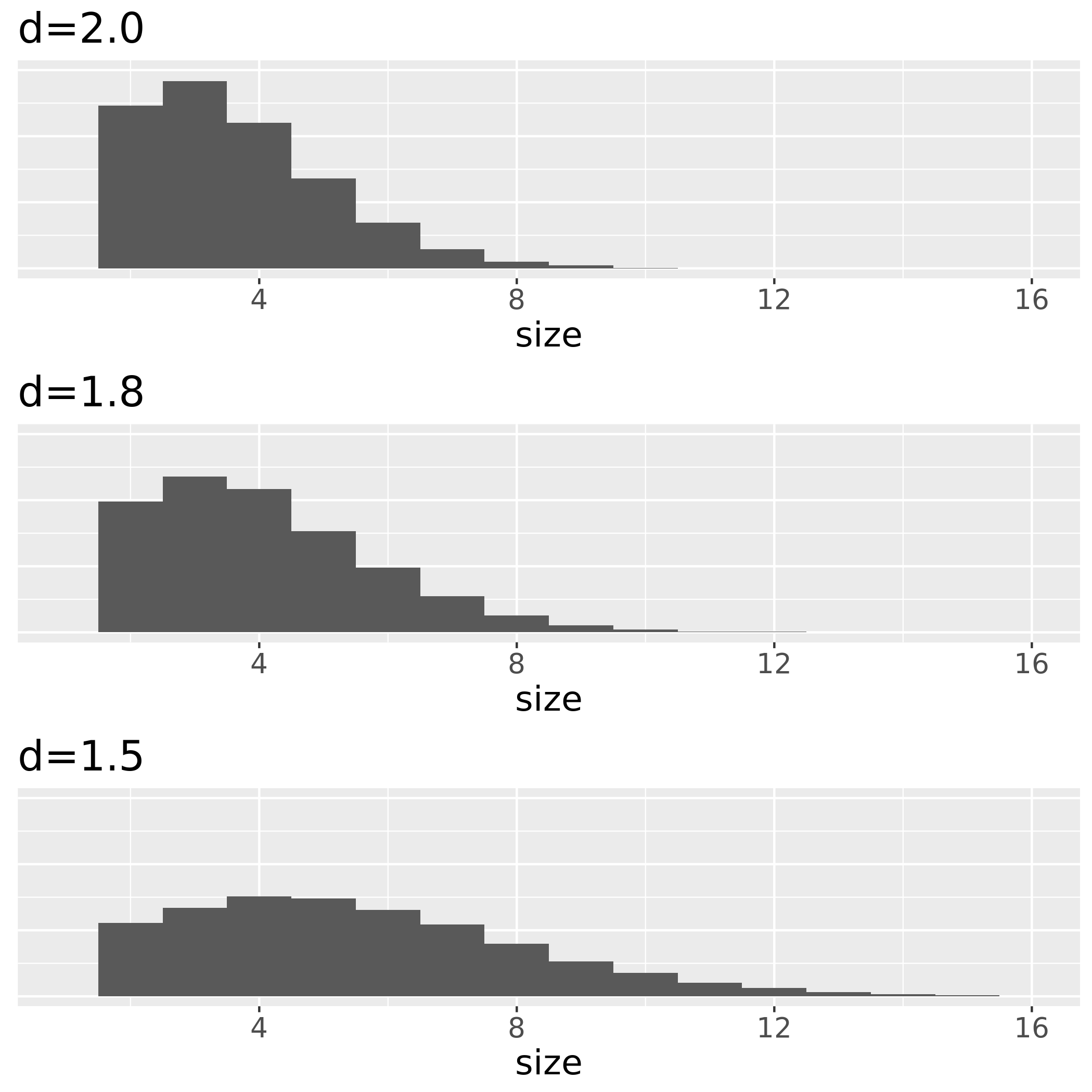}
    \caption{Distribution of expression sizes when varying the value of the decay parameter $d$. The size of an expression is measured as the number of its digit nodes.}
    \label{fig:size_dist}
\end{figure}

We implement the neural models in PyTorch 1.0.0. We use the following
model architecture: encoder embedding layer size 64, encoder LSTM size
128, for the regression models, MLP with 1 hidden layer of size 256; for the
sequence-to-sequence model the decoder hyper-parameters are the same
as the encoder. The symbols are predicted via a linear projection
layer from hidden state, followed by a softmax. Training proceeds
following a curriculum: we first train on 100,000 batches of size 32
of random expressions sampled with decay $d=2.0$, followed by 200,000
batches with $d=1.8$ and finally 400,000 batches with $d=1.5$. We
optimize with Adam with learning rate $0.001$. We report results on
expressions sampled with $d=1.5$. See Figure~\ref{fig:size_dist} for
the distribution of expression sizes for these values of $d$.

We report all results for two conditions: randomly initialized, and
trained, in order to quantify the effect of learning on the activation
patterns. 
The trained model is chosen by saving model weights
during training every 10,000 batches and selecting the weights with
the smallest loss on 1,000 held-out validation expressions. Results
are reported on separate test data consisting of 2,000 expressions and
200 reference expressions for RSA$_{\textsc{regress}}$ embedding.

\subsection{Results}
\label{sec:synthetic:results}

\begin{table*}
    \centering
    {\small
    \begin{tabular}{l r rr rrrr rrrr}
    & \multicolumn{1}{c}{} & \multicolumn{2}{c}{Diagnostic} & \multicolumn{4}{c}{RSA} & \multicolumn{4}{c}{RSA$_{\textsc{regress}}$}\\ \cmidrule(lr){3-4}\cmidrule(lr){5-8}\cmidrule(lr){9-12}
    Encoder                 & Loss    & Value    & Depth    & Value    & Depth    & TK($1$)        & TK($0.5$)           & Value    & Depth    & TK($1$)          & TK($0.5$) \\\toprule
\sc Random                  &         & 0.01     & 0.80     & 0.01     & 0.23     & -0.24          & -0.33               & -0.01    & 0.57     & 0.41             &  0.63            \\
\sc Semantic eval.          & 0.07    & \bf 0.97 & 0.70     & \bf 0.62 & 0.05     & 0.02           & 0.01                & \bf 0.97 & 0.55     & 0.38             &  0.61            \\
\sc Tree Depth              & 0.00    & -0.03    & \bf 1.00 & 0.01     & \bf 0.72 & 0.10           & -0.06               & -0.03    & \bf 0.97 & 0.49             &  0.87            \\
\sc Infix-to-Prefix         & 0.00    & 0.02     & 0.97     & -0.00    & 0.64     & \bf 0.35       & \bf 0.53            & 0.02     & 0.88     &\bf  0.58         &  \bf 0.96        \\
    \end{tabular}
    }
    \caption{Scores for diagnostic regression, RSA, and
      RSA$_{\textsc{regress}}$ with respect to expression value,
      expression tree depth and the Tree Kernel (TK) with $\lambda=1$ and $\lambda=0.5$. All scores are Pearson's correlation coefficients. For the diagnostic
      model and RSA$_{\textsc{regress}}$ they are cross-validated correlations between target and predicted values.
      The randomly initialized encoder is the same for
      all encoder types, and thus there
      is only a single row for the {\sc Random} encoder. The loss
      column shows the loss of the full model on the test data: mean
      squared error for {\sc Semantic evaluation} and {\sc Tree
        depth}, and cross-entropy for {\sc Infix-to-prefix}.}
    \label{tab:results_main}
\end{table*}

Table~\ref{tab:results_main} shows the results of our experiments,
where each row shows a different encoder type and each column a
different target task. 

\paragraph{Semantic value and tree depth}
As a first sanity check, we would like to see whether the RSA
techniques show the same pattern captured by the diagnostic models.
As expected, both diagnostic and RSA scores are the highest when the
objective function used to train the encoder and the analytical
reference representations match: for example, the {\sc semantic
  evaluation} encoder scores high on the {\it semantic value}
reference, both for the diagnostic model and the RSA. Furthermore, the
scores for the value and depth reference representation according to
the diagnostic model and according to RSA$_{\textsc{regress}}$ are in
agreement. The scores according to RSA in some cases show a different
picture. This is expected, as RSA answers a substantially different
question than the other two approaches: it looks at how the whole
representations match in their similarity structure, whereas both the
diagnostic model and RSA$_{\textsc{regress}}$ focus on the part of the
representation that encodes the target information the strongest.

\paragraph{Tree Kernel}

We can use both RSA and RSA$_{\textsc{regress}}$ for exploring whether
the hidden activations encode any structural representation of syntax:
this is evident in the scores yielded by the TK reference
representations. As expected, the highest scores for both methods are
gained when using {\sc Infix-to-prefix} encodings, the task that
relies the most on the hierarchical structure of an input string.
RSA$_{\textsc{regress}}$ yields the second-highest score for {\sc Tree
  depth} encodings, which also depend on aspects of tree
structure. The overall pattern for the TK with different values of the
discounting parameter $\lambda$ is similar, even though the absolute values of
the scores vary.
What is unexpected is the results for the random encoder,
which we turn to next.

\paragraph{Random encoders}
The non-random nature of the activation patterns of randomly
initialized models \cite[e.g.,][]{zhang2018language} is also strongly in evidence in our
results. For example the random encoder has quite a high score for
diagnostic regression on {\it tree depth}.  Even more striking is the fact
that the random encoder has substantial {\it negative } RSA score for
the Tree Kernel: thus, expression pairs more similar
according to the Tree Kernel are less similar according to the random
encoder, and vice-versa.

\begin{figure}
    \centering
    \includegraphics[scale=0.3]{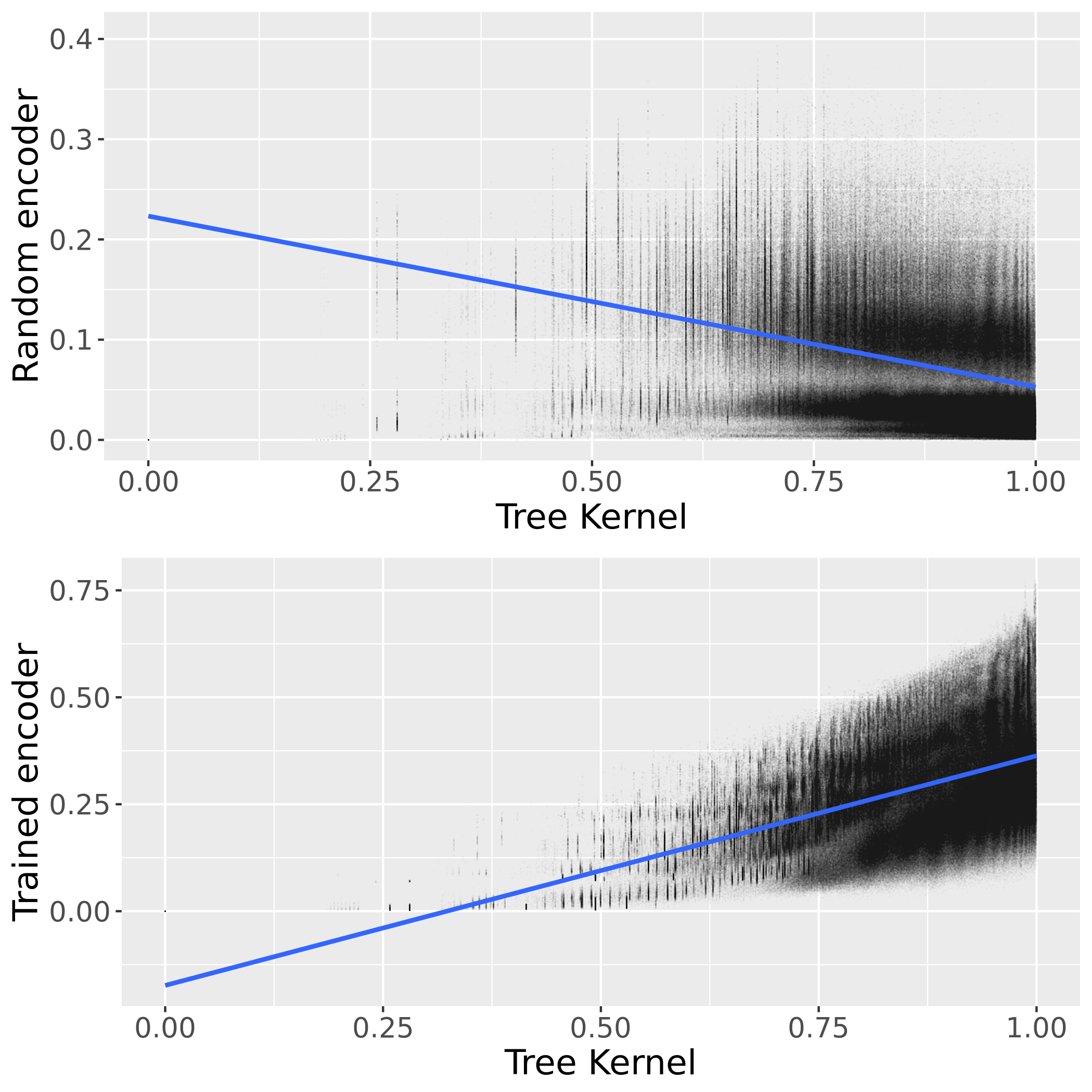}
    \caption{Scatterplot of dissimilarity values according to random
      encoder or trained {\sc Infix-to-Prefix} encoder and the Tree
      Kernel ($\lambda = 0.5$)}
    \label{fig:scatter_rsa_prefix}
\end{figure}

When applying RSA we can inspect the full correlation pattern via a
scatter-plot of the dissimilarities in the reference and encoder
representations. Figure~\ref{fig:scatter_rsa_prefix} shows the data
for the random encoder and the Tree Kernel representations. As can be
seen, the negative correlation for the random encoder is due to the
fact that according to the Tree Kernel, expression pairs tend to have
high dissimilarities, while according to the random encoder's
activations they tend to have overall low dissimilarities. For the
trained {\sc Infix-to-prefix} encoder the dissimilarities are clearly
positively correlated with the TK dissimilarities.

Thus the raw correlation value for the trained encoder is a biased
estimate of the effect of learning, as learning has to overcome
the initially substantial negative correlation: a better estimate is
the difference between scores for the learned and random model. It is
worth noting that the same approach would be less informative for the
diagnostic model approach or for RSA$_{\textsc{regress}}$. For a regression model
the correlation scores will be positive, and when taking the
difference between learned and random scores, they may cancel out,
even though a particular information may be predictable from the
random activations in a completely different way than from the learned
activations. This is what we see for the RSA$_{\textsc{regress}}$
scores for random vs.\ {\sc infix-to-prefix} encoder: the scores
partially cancel out, and given the pattern in
Figure~\ref{fig:scatter_rsa_prefix} it is clear that subtracting them
is misleading. It is thus a good idea to complement the
RSA$_{\textsc{regress}}$ score with the plain RSA correlation score in
order to obtain a full picture of how learning affects the neural
representations.

\vspace{.2cm}
Overall, these results show that RSA$_{\textsc{regress}}$ can be used to answer
the same sort of questions as the diagnostic model.  It has the added
advantage of being also easily applicable to structured symbolic
representations, while the RSA scores and the full RSA correlation
pattern provides a complementary source of insight into neural
representations. Encouraged by these findings, we next apply both RSA
and RSA$_{\textsc{regress}}$ to representations of natural language
sentences.

\section{Natural language}
\label{sec:natural}

Here we use our proposed RSA-based techniques to compare
tree-structure representations of natural language sentences with
their neural representations captured by sentence embeddings.  Such
embeddings are often provided by NLP systems trained on unlabeled text,
using variants of a language modeling objective \citep[e.g.][]{peters2018deep}, next and previous
sentence prediction \citep{kiros2015skip,logeswaran2018efficient}, or
discourse based objectives \citep{nie2017dissent,jernite2017discourse}. Alternatively
they can be either fully trained or fine-tuned on annotated data using
a task such as natural language inference
\citep{conneau-etal-2017-supervised}. In our experiments we use one of
each type of encoders.

\subsection{Encoders}
\paragraph{Bag of words}
As a baseline we use a classic bag of words model where a sentence is
represented by a vector of word counts. We do not exclude any words
and use raw, unweighted word counts.

\vspace{-.2cm}
\paragraph{Infersent}
This is the supervised model described in
\citet{conneau-etal-2017-supervised} based on a bidirectional LSTM
trained on natural language inference. We use the {\tt infersent2} model
with pre-trained fastText \citep{bojanowski2017enriching}
word embeddings.\footnote{Available at
  \href{https://github.com/facebookresearch/InferSent}{https://github.com/facebookresearch/InferSent}.}
We also test a randomly initialized version of this model, including
random word embeddings.

\vspace{-.2cm}
\paragraph{BERT}
This is an unsupervised model based on the Transformer architecture
\citep{vaswani2017attention} trained on a cloze-task and next-sentence
prediction \citep{DBLP:journals/corr/abs-1810-04805}. We use the
Pytorch version of the large 24-layer model ({\tt
  bert-large-uncased}).\footnote{Available at
  \href{https://github.com/huggingface/pytorch-pretrained-BERT}{https://github.com/huggingface/pytorch-pretrained-BERT}.}
We also test a randomly initialized version of this model.

\subsection{Experimental settings}
\label{sec:natural:experimental}
\paragraph{Data}
We use a sample of data from the English Web Treebank (EWT)
\citep{ewt} which contains a mix of English weblogs, newsgroups,
email, reviews and question-answers manually annotated for syntactic
constituency structure. We use the 2,002 sentences corresponding to
the development section of the EWT Universal Dependencies
\citep{silveira14gold}, plus 200 sentences from the training section
as reference sentences when fitting RSA$_{\textsc{regress}}$.

\paragraph{Tree Kernel}
Prior to computing the Tree Kernel scores we delexicalize the constituency trees by replacing all terminals (i.e.\ words) with a single placeholder value {\tt X}. This ensures that only syntactic structure, and not lexical overlap, contributes to kernel scores.
We compute kernels for the values of $\lambda \in \{1, \frac{1}{2}\}$. 

\vspace{-.2cm}
\paragraph{Embeddings}
For the BERT embeddings we use the vector associated with the first
token ({\tt CLS}) for a given layer.  For Infersent, we use the
default max-pooled representation.

\vspace{-.2cm}
\paragraph{Fitting}
When fitting RSA$_{\textsc{regress}}$ we use L2-penalized multivariate
linear regression. We report the results for the value of the
$\textrm{penalty}=10^n$, for $n \in \{-3,-2,-1,0,1, 2\}$, with the
highest $10$-fold cross-validated Pearson's $r$ between target and
predicted similarity-embedded vectors.

\subsection{Results}
\label{sec:natural:results}
\begin{table}
    \centering
    {\small
      \begin{tabular}{lcrrr}
Encoder    & Train & $\lambda$ & RSA & RSA$_{\textsc{regress}}$\\\toprule        
BoW        &       &  0.5  &  0.18  &  0.50  \\
Infersent  &  $-$  &  0.5  &  0.24  &  0.51  \\
BERT last  &  $-$  &  0.5  &  0.12  &  0.49  \\
BERT best  &  $-$  &  0.5  &  0.14  &  0.53  \\
Infersent  &  $+$  &  0.5  &  0.30  &  \bf 0.71  \\
BERT last  &  $+$  &  0.5  &  0.16  &  0.59  \\
BERT best  &  $+$  &  0.5  &  \bf 0.32  &  0.70  \\\midrule
BoW        &          &  1.0  &  -0.01  &  0.40  \\
Infersent  &  $-$  &  1.0  &  0.00  &  0.48  \\
BERT last  &  $-$  &  1.0  &  -0.08  &  0.50  \\
BERT best  &  $-$  &  1.0  &  -0.07  &  0.52  \\
Infersent  &  $+$  &  1.0  &  0.10  &  0.59  \\
BERT last  &  $+$  &  1.0  &  0.03  &  0.53  \\
BERT best  &  $+$  &  1.0  &  \bf 0.18  &  \bf 0.60  \\
\end{tabular}
    
    }
    \caption{Correlation scores for encoders against Tree Kernel with
      varying $\lambda$. Scores for both RSA and
      RSA$_{\textsc{regress}}$ are Pearson's r.
      The column {\it Train} indicates whether the encoder
      (including the word embeddings) is randomly initialized ($-$),
      or trained ($+$). For BERT, we report 
      scores for the topmost (last) layer and for the layer which maximizes
      the given score (best).}
    \label{tab:rsa-natural}
\end{table}

Table~\ref{tab:rsa-natural} shows the results of applying RSA and
RSA$_{\textsc{regress}}$ on five different sentence encoders, using
the Tree Kernel reference. Results are reported using two different
values for the Tree Kernel parameter $\lambda$. 

As can be seen, with $\lambda=\frac{1}{2}$, all the encoders show a substantial
RSA correlation with the parse trees. The highest scores are
achieved by the trained Infersent and BERT, but even Bag of
Words and untrained versions of Infersent and BERT show a sizeable
correlation with syntactic trees according to both RSA and
RSA$_{\textsc{regress}}$.

When structure matching is strict ($\lambda=1$), only trained
BERT and Infersent capture syntactic information according to RSA;
however, RSA$_{\textsc{regress}}$ still shows moderate correlation for
BoW and the untrained versions of BERT and Infersent. Thus
RSA$_{\textsc{regress}}$ is less sensitive to the value of $\lambda$
than RSA since changing it from $\frac{1}{2}$ to $1$ does not alter results in a
qualitative sense.

\begin{figure}
    \centering
    \includegraphics[scale=0.45]{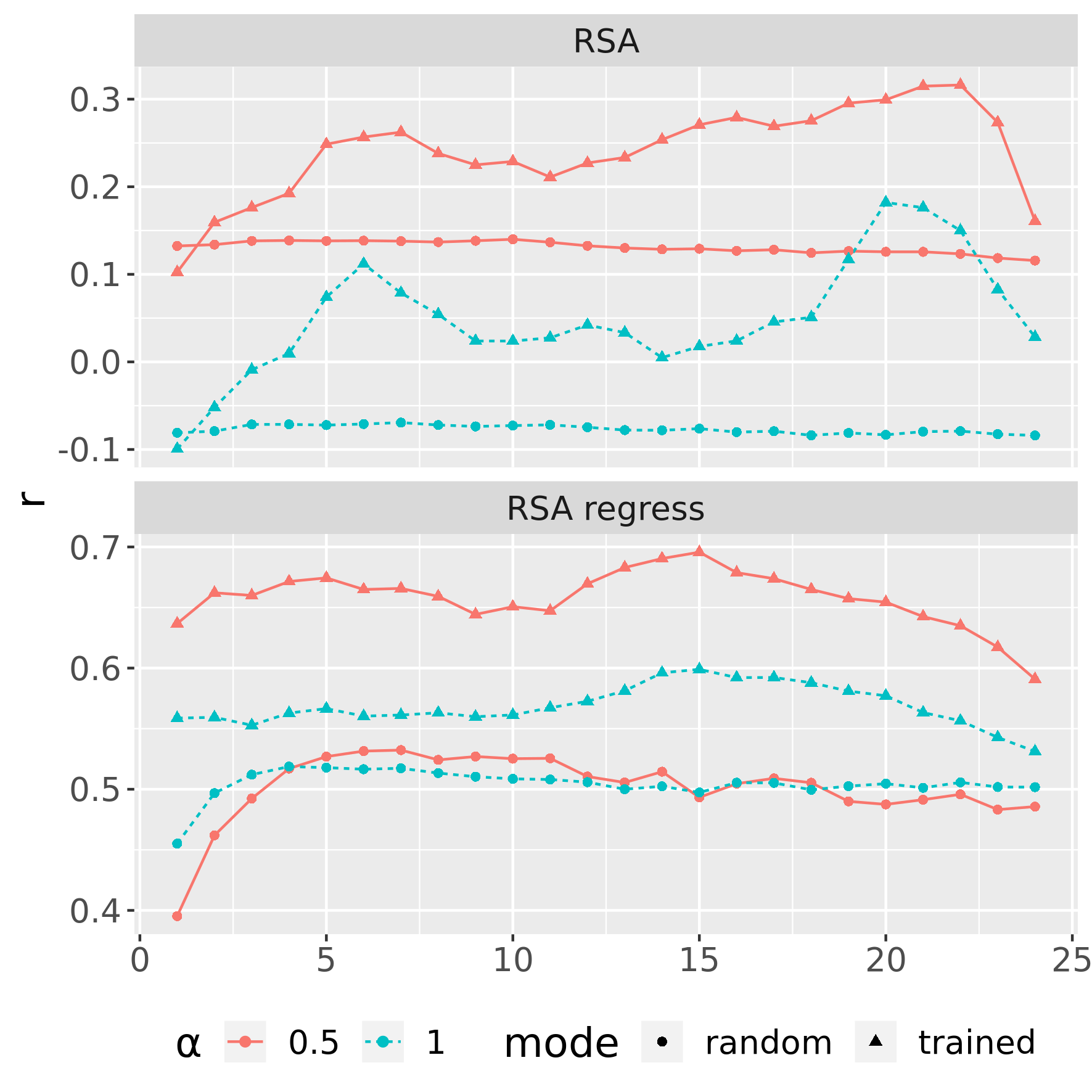}
    \caption{RSA and RSA$_{\textsc{regress}}$ scores for embeddings from all
      the layers of BERT vs Tree Kernel for two values of
      $\lambda$. Both randomly initialized and trained versions of BERT
      are shown. The embeddings are vectors at the first token ({\tt
        CLS}) at each layer.}
    \label{fig:bert-layers}
\end{figure}

Figure~\ref{fig:bert-layers} shows how RSA and
RSA$_{\textsc{regress}}$ scores change when correlating Tree Kernel
estimates with embeddings from different layers of BERT. For trained
models, scores peak between layers 15--22 (depending on metric
and $\lambda$) and decline thereafter, which indicates that
the final layers are increasingly dedicated to encoding aspects of
sentences other than pure syntax.

\section{Conclusion}
\label{sec:conclusion}
We present two RSA-based methods for correlating neural and syntactic representations of
language, using tree kernels as a measure of similarity between
syntactic trees. Our results on arithmetic expressions confirm that both
versions of structured RSA capture correlations between
different representation spaces, while providing
complementary insights.  We apply the same techniques to English
sentence embeddings, and show where and to what extent each
representation encodes syntactic information.
The proposed methods are general and applicable not just to
constituency trees, but given a similarity metric, to any symbolic
representation of linguistic structures including dependency trees or
Abstract Meaning Representations. We plan to explore these options in
future work. A toolkit with the implementation of our methods is available at \href{https://github.com/gchrupala/ursa}{https://github.com/gchrupala/ursa}.

\bibliography{biblio}
\bibliographystyle{acl_natbib}

\end{document}